\documentclass{article} 
\usepackage{arxiv}

\usepackage[utf8]{inputenc} 
\usepackage[T1]{fontenc}    
\usepackage{hyperref}       
\usepackage{url}            
\usepackage{booktabs}       
\usepackage{amsfonts}       
\usepackage{nicefrac}       
\usepackage{microtype}      
\usepackage{lipsum}
\usepackage{graphicx}
\graphicspath{ {./images/} }

\usepackage{microtype}
\usepackage{times}
\usepackage{soul}
\usepackage{url}
\usepackage[utf8]{inputenc}
\usepackage[small]{caption}
\usepackage{graphicx}
\usepackage{amsmath}
\usepackage{amsthm}
\usepackage{booktabs}
\usepackage{algorithm}
\usepackage{algorithmic}
\usepackage[shortlabels]{enumitem}
\usepackage[TABBOTCAP]{subfigure}
\usepackage{multirow}
\usepackage{diagbox}

\usepackage{multirow}

\usepackage{color}

\def\EM{{\mathcal E}}

\def\OM{{\mathcal O}}
\def\PM{{\mathcal P}}

\def\WM{{\mathcal W}}

\def\e{{\bf e}}

\def\o{{\bf o}}

\def\w{{\bf w}}

\def\0{{\bf 0}}
\def\1{{\bf 1}}

\usepackage{amsmath}

\title{Learning Contextual Causality from Time-consecutive Images}

\author{
  Hongming Zhang$^1$, Yintong Huo$^{1,2}$\thanks{This work was done when the author was visiting HKUST.}, $\,$ Xinran Zhao$^1$, Yangqiu Song$^1$, Dan Roth$^3$\\
  $^1$Department of Computer Science and Engineering, HKUST\\
  $^2$Department of Computer Science and Engineering, CUHK\\
  $^3$Department of Computer and Information Science, UPenn\\
  \texttt{hzhangal@cse.ust.hk, ythuo@cse.cuhk.edu.hk, xzhaoar@connect.ust.hk,}\\ \texttt{yqsong@cse.ust.hk, danroth@seas.upenn.edu} \\
}
\date{}

\begin{document}
\maketitle
\begin{abstract}

Causality knowledge is crucial for many artificial intelligence systems.
Conventional textual-based causality knowledge acquisition methods typically require laborious and expensive human annotations.
As a result, their scale is often limited.
Moreover, as no context is provided during the annotation, the resulting causality knowledge records (e.g., ConceptNet) typically do not take the context into consideration.
To explore a more scalable way of acquiring causality knowledge, in this paper, we jump out of the textual domain and investigate the possibility of learning contextual causality from the visual signal.
Compared with pure text-based approaches, learning causality from the visual signal has the following advantages: (1) Causality knowledge belongs to the commonsense knowledge, which is rarely expressed in the text but rich in videos; (2) Most events in the video are naturally time-ordered, which provides a rich resource for us to mine causality knowledge from; (3) All the objects in the video can be used as context to study the contextual property of causal relations.
In detail, we first propose a high-quality dataset Vis-Causal and then conduct experiments to demonstrate that with good language and visual representation models as well as enough training signals, it is possible to automatically discover meaningful causal knowledge from the videos.
Further analysis also shows that the contextual property of causal relations indeed exists, taking which into consideration might be crucial if we want to use the causality knowledge in real applications, and the visual signal could serve as a good resource for learning such contextual causality.
Vis-Causal and all used codes are available at: \url{https://github.com/HKUST-KnowComp/Vis_Causal}.


\end{abstract}
\section{Introduction}\label{sec-introduction}



Humans possess a basic knowledge about facts and understandings for commonsense of causality in our everyday life. 
For example, if we leave five minutes late, we will be late for the bus; if the sun is out, it's not likely to rain; and if we are hungry, we need to eat.
Such causality knowledge has been shown to be helpful for many NLP tasks~\cite{DBLP:conf/acl/OhTHSSO13,DBLP:conf/acl/HashimotoTKSVOK14,DBLP:conf/acl/RothWNF18}.
Thus, it is valuable to teach machines to understand causality~\cite{pearl2018book}.

Causal relations in the commonsense domain are typically contributory and contextual~\cite{bunge2017causality}. 
By contributory\footnote{The other two levels are absolute causality (the cause is necessary and sufficient for the effect) and conditional causality (the cause is necessary but not sufficient for the effect), which commonly appear in the scientific domain rather than our daily life.}, we mean that the cause is neither necessary nor sufficient for the effect, but it strongly contributes to the effect. 
By contextual, we mean that some causal relations only make sense in a certain context.
The contextual property of causal relations is important for both the acquisition and application of causal knowledge.
For example, if some people tell the AI assistant (e.g. Siri) ``they are hungry'' in a meeting, a basic assistant may suggest them to order food because it has the knowledge that `being hungry' causes `eat food'. A better assistant may suggest ordering food \textbf{after} the meeting because it knows that the causal relation between `being hungry' and `eat food' may not be plausible in the meeting context. 
Without understanding the contextual property of causal knowledge, achieving such a level of intelligence would be challenging.

To help machines better understand the causality commonsense, many efforts have been devoted into developing the causality knowledge bases. 
For example, ConceptNet~\cite{liu2004conceptnet} and ATOMIC~\cite{DBLP:conf/aaai/SapBABLRRSC19} leverage 
human-annotation to acquire small-scale but high-quality causality knowledge.
After that, people try to leverage linguistic patterns (e.g., two events connected with ``\textit{Because}'')~\cite{DBLP:conf/acl/HideyM16,DBLP:conf/kr/LuoSZHW16,ASER2020} to acquire causality knowledge from textual corpus.
However, causality knowledge, especially those trivial knowledge for humans, are rarely formally expressed in documents~\cite{liu2004conceptnet}, a pure text-based approach might struggle at covering all causality knowledge.
Besides that, none of them take the aforementioned contextual property of causal knowledge into consideration, which may restrict their usage in downstream tasks.

In this paper, we propose to ground causality knowledge into the real world and explore the possibility of acquiring causality knowledge from visual signals (i.e., images in time sequence, which are cropped from videos). 
By doing so, we have three major advantages: (1) Videos can be easily acquired and can cover rich commonsense knowledge that may not be mentioned in the textual corpus; (2) Events contained in videos are naturally ordered by time. As discussed by~\cite{DBLP:conf/acl/RothWNF18}, there exists a strong correlation between temporal and causal relations, and thus such time-consecutive images can become a dense causality knowledge resource; (3) Objects from the visual signals can act as the context for detected causality knowledge, which can remedy the aforementioned lack of contextual property issue of existing approaches. 
\begin{figure}
    \centering
    \includegraphics[width=0.6\linewidth]{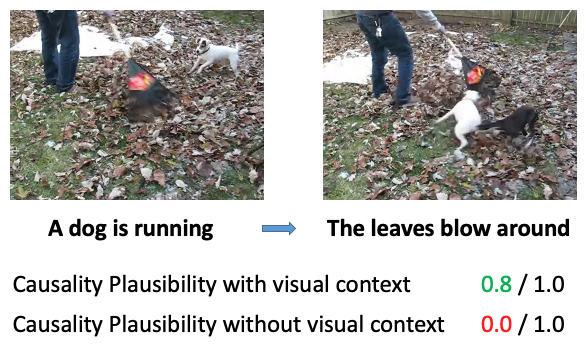}
    \caption{An example of Vis-Causal. Based on the annotated plausibility, there exists a strong causal relation from `A dog is running' to `The leaves blow around' with the visual signal as context (i.e., on leaves), but that causal relation is no longer plausible without the context.}

    \label{fig:dataset-demo}
\end{figure}
To be more specific, we first define the task of mining causality knowledge from time-consecutive images and propose a high-quality dataset (Vis-Causal). 
To study the contextual property of causal relations, for each pair of events, we provide two kinds of causality annotations: one is the causality given certain context and the other one is the causality without context. 
Distribution analysis and case studies are conducted to analyze the contextual property of causality.
An example from Vis-Causal is shown in Figure~\ref{fig:dataset-demo}, where the causal relation between ``dog is running'' and ``blowing leaves'' only makes sense when the context is provided because the dog is running on the leaves, so its high speed and quickly-moved pow cause the leaves blow around. Without the context  ``leaves on the ground'', this causal relation is implausible.
After that, we propose a Vision-Contextual Causal (VCC) model, which can effectively leverage both the pre-trained textual representation and visual context to acquire causality knowledge and can be used as a baseline method for future works.
Experimental results demonstrate that even though the task is still challenging, by jointly leveraging the visual and contextual representation, the proposed model can better identify meaningful causal relations from time-consecutive images.
To summarize, the contributions of this paper are three-fold: (1) We formally define the task of mining contextual causality from the visual signal; (2) We present a high-quality dataset Vis-Causal; (3) We propose a Vision-Contextual Causal (VCC) model to demonstrate the possibility of mining contextual causality from the vision signal.

The rest of the paper is organized as follows. In Section 2, we formally define the task of learning contextual causality from the visual signal. After that, we present the construction details about Vis-Causal in Section 3. 
In Section 4 and 5, we present the details about the proposed VCC model and the experiments.
In the end, we use Section 6 to introduce related works and Section 7 to conclude this paper.

\section{The Task Definition}\label{sec:task-definition}

As introduced in the introduction, the ultimate goal of this work is to acquire contextual causality knowledge from videos. However, as current models cannot afford processing videos directly, we simplify the task into mining causality knowledge from time-consecutive images, which are cropped from the video. 
Thus, we formally define the task as follows. Each image pair $P \in \PM$, where $\PM$ is the overall image pair set, consists of two images $I_1$ and $I_2$, sampled from same video, in a temporal order (i.e., $I_1$ appears before $I_2$). For each $P$, our goal is to identify all possible causal relations between the contained images. Normally, this task contains two sub-tasks: identifying events in images and identifying causality relation between contained events. As there exists a huge overlap between the event identification task and the scene graph 
generation task~\cite{DBLP:conf/cvpr/XuZCF17} in the computer vision (CV) community, which has been extensively studied
~\cite{DBLP:conf/eccv/YangLLBP18,DBLP:conf/eccv/LiOZSZW18}, in this work, we focus on the second sub-task. We assume that the event sets contained in $I_1$ is denoted as $\EM_1$ and the event sets contained in all images sampled from $V_1$ is denoted as $\EM_v$. For each event $e_1 \in \EM_1$, our goal is finding all events $e_2 \in \EM_v$ such that $e_1$ causes $e_2$.



\section{The Vis-Causal Dataset}\label{sec:dataset}

In this section, we introduce the details about the creation of Vis-Causal, which was carried out in four steps: (1) Pre-processing the raw video data into frames for further annotation; (2) Identifying the contained events from the frames; (3) First-round of causal annotation, which requires annotators to write down events that can be caused by the given event; (4) Second-round of causal annotation, which refines the quality of the annotation from step three via more fine-grained annotation and two settings are included (one is with the visual context and the other one is without any context).
We select Amazon MTurk\footnote{https://www.mturk.com/}
as the annotation platform and show survey examples in the appendix.
We elaborate on each step as follows.

\subsection{Data Source}

In order to acquire a broad coverage of daily life causality, we chose to use ActivityNet~\cite{DBLP:conf/cvpr/HeilbronEGN15}, which contains short videos from YouTube, as the video resource. 
We randomly selected 1,000 videos. 
For each video,  we took five uniformly sampled screen-shots and took adjoined screen-shots as pairs of time-consecutive images to better capture the chained events.
As a result, we collected 4,000 image pairs.

\subsection{Event Identification}\label{sec:event-identification}
In the first step, for each image pair, we invited three annotators to write down any events they can identify in the first image. Clear instructions and examples were given to help annotators understand the definition of events and our task. 
We invited three annotators for each image, resulting in 12,000 events in total for 4,000 image pairs.


\subsection{First-round Causal Annotation}\label{sec:candidate-preparation}

After identifying events, we invited annotators to identify related event pairs, which are used as candidates for next-step causal relation annotation.
For each pair of time-consecutive images, we selected all three identified events in the first image, and for each one of them, we asked annotators to describe one event that happens in the second image and is caused by the selected event, which happens in the first image. 
If no suitable event was found, which is possible due to the sparseness of causal relations, the annotators could choose `None' as the answer.
For each question, we invited three different annotators to provide annotations.
After filtering out answers that contain `None' or has less than two words, we obtained 23,558 event pair candidates.
On average, we kept 5.89 candidate event pairs for each image pair. 




\subsection{Second-round Causal Annotation}\label{sec:causal-annotation}

As crowd-workers, unlike experts, are less likely to effectively distinguish the difference between inference and causality (For example, `A cat is running' infers `An animal is running' but `A cat is running' doesn't cause `an animal is running'), such relations are often mistakenly annotated as causal relation in the first-round annotation based on our observation.
To remedy this problem, in the second round of annotation, we first present a clear definition and several examples of all possible relations (e.g., `Inference' and `Causality') and then ask annotators to select the most plausible relations for all candidate event pairs, which is more fine-grained and thus achieves the better annotation quality.
Besides that, to investigate the contextual property of causality knowledge, two settings are considered for the annotation (one with the context and one without).
For each setting, we invite five annotators for annotating each event pair.
Following previous works~\cite{DBLP:conf/emnlp/ReisingerM10,DBLP:journals/coling/HillRK15}, we employ Inter Annotator Agreement (IAA), which computes the average agreement of an annotator
with the average of all other annotators, to evaluate the overall annotation quality.
As a result, we achieve 78\% and 76\% IAA scores for ``\textit{with context}'' and ``\textit{without context}'' settings respectively.
We achieve slightly lower agreement in the ``without context'' setting.
One possible explanation is that when no context is provided, different people may think about different contexts and thus their annotations about causal relations could be slightly different.

\subsection{Annotation Analysis}


\begin{figure*}
    \centering
    \includegraphics[width=\linewidth]{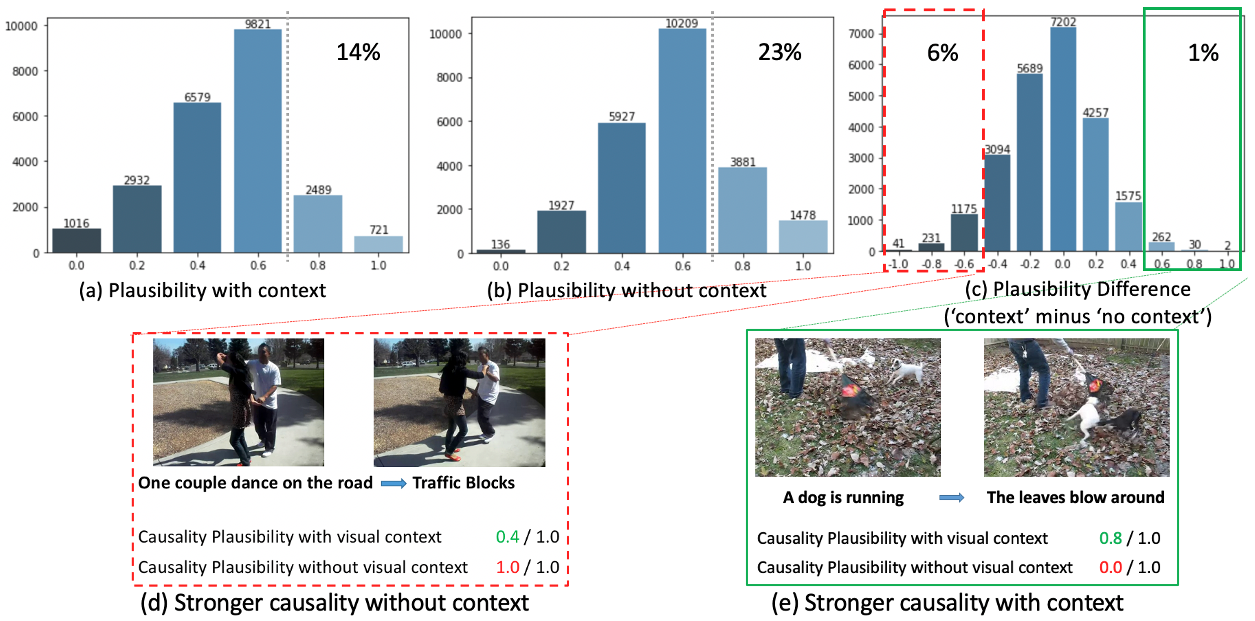}
    \caption{Distribution of plausibility scores under different settings and their difference.}
    \label{fig:distribution}
\end{figure*}

\begin{table}[t]
\small
    \centering
    \begin{tabular}{c|c|c|c|c|c}
    \toprule
      & \#Videos & \#Images & \#Pairs & \#Positives & \#Cand. \\
         \midrule
        Train & 800 & 4000 & 3200 & 2599 & 31.8 \\
        Dev & 100 & 500 & 400 & 329 & 32.1\\
        Test & 100 & 500 & 400 & 282& 32.2\\
    \bottomrule
    \end{tabular}
    \vspace{0.1in}
    \caption{Dataset statistics. \textbf{\#Videos}, \textbf{\#Images} and \textbf{\#Pairs} denotes the number of videos, images and image pairs, respectively. \textbf{\#Positives} denotes the number of event pairs labeled with causality relation. \textbf{\#Cand.} denotes the average length of the candidate event list.}
    \label{tab:stats}
\end{table}

The distribution of annotation results for both settings are shown in Figure~\ref{fig:distribution}(a) and Figure~\ref{fig:distribution}(b) respectively.
For each pair of events, we compute the plausibility based on voting. 
For example, if four out of five annotators vote ``causal'', its causal plausibility is 0.8.
In general, we can see that for both settings, the majority of the candidate events pairs have weak causal relations and only a small portion of the candidates contain strong causal relations, especially for the ``\textit{with context}'' setting.
One possible explanation is that when no context is provided, humans can think about multiple contexts and find the most suitable one such that the causal relation is plausible in that scenario.
However, when the visual context is provided, where the scenario is fixed, humans do not have the freedom to choose a suitable scenario by themselves.
As a result, Annotators tend to annotate more plausible causal relations in the ``\textit{no context}'' setting.

To investigate the contextual property of causal relations, we show the distribution of plausibility difference (``\textit{with context}'' minus ``\textit{without context}'') in Figure~\ref{fig:distribution}(c).
From the result, we can observe that about 6\% of event pairs, which is indicated with the dashed box, have stronger causal relations without any context, while about 1\% of event pairs, which is indicated with the solid box, have a stronger causal relation when the visual context is provided.
Two examples of both cases are shown in Figure~\ref{fig:distribution}(d) and \ref{fig:distribution}(e) respectively.


\subsection{Dataset Splitting and Statistics}

We split the dataset into train, dev, and test sets based on the original split of ActivityNet~\cite{DBLP:conf/cvpr/HeilbronEGN15} and collect 800, 100, and 100 videos for the train, dev, and test set respectively.
We select positive causal relations based on the annotation under the ``\textit{with context}'' setting.
If at least four of five annotators think there exists a causal relation between a pair of events given the context, we will treat it as a positive example. As a result, we got 2,599, 329, and 282 positive causal pairs for the train, dev, and test set, respectively. 
On average, each event pair contains 11.41 words and the total vocabulary size is 10,566.
We summarize the detailed dataset statistics in Table~\ref{tab:stats}.

\section{The VCC Model}\label{sec:model}

\begin{figure}
    \centering
    \includegraphics[width=0.6\linewidth]{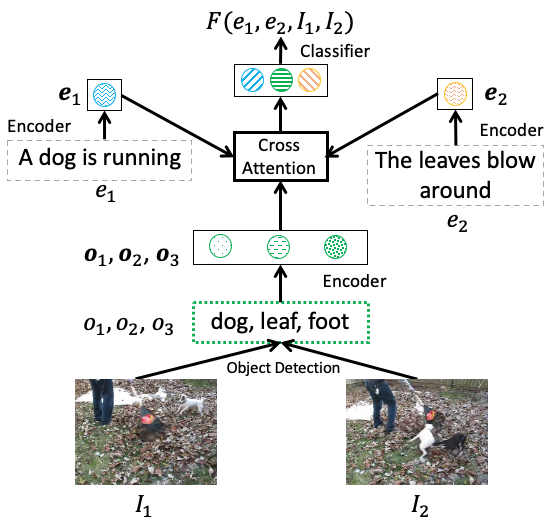}
    \caption{Demonstration of the proposed model. We use $\e_1$ and $\e_2$ to denote the vector representation of $e_1$ and $e_2$ after encoding respectively. Assuming that three objects ($o_1$, $o_2$, and $o_3$), whose vector representations are denoted as $\o_1$, $\o_2$, and $\o_3$ respectively, are extracted from the two context images, a cross attention module is proposed to jointly leverage the event and context representation to make the final prediction.}
    \label{fig:our-model}
\end{figure}

In this section, we introduce the proposed Vision-Contextual Causal (VCC) Model, which leverages both the visual context and contextual representation of events to predict the causal relations, and we show the overall framework in Figure~\ref{fig:our-model}.
In total, we have three major components: event encoding, which encodes the two events into vectors for further prediction; visual context encoding, which encodes the context frame such that the context can be utilized in the model; and cross attention, which aims at finding the best context and event representation via the attention mechanism~\cite{DBLP:conf/nips/VaswaniSPUJGKP17}. 
The details about these components are introduced as follows.

\subsection{Textual Event Encoding}

As both $e_1$ and $e_2$ are represented with natural language, we begin with converting them into vector representations.
In this work, we leverage a pre-trained language representation model BERT~\cite{DBLP:conf/naacl/DevlinCLT19} to encode all events. Assuming that after the tokenization, event $e$ contains n tokens $w_1, w_2, ..., w_n$, 
we denote their contextualized representations after BERT as $\w_1, \w_2, ..., \w_n$.

\subsection{Visual Context Encoding}

Following the common approach in multi-modal approaches approaches~\cite{DBLP:conf/cvpr/XuZCF17,DBLP:conf/cvpr/DasKGSYMPB17} we first leverage an object detection module to detect objects from images and use all extracted objects to represent the visual context.
Assuming that for $I_1$ and $I_2$, we extract $m_1$ and $m_2$ objects respectively. After combining all objects from two images together and sorting them based on the confidence score provided by the object detection module, we keep the top $m$ objects and denote them as $o_1, o_2, ..., o_m$. The motivation for that operation is to avoid the influence of noise introduced by the object detection module. As all objects are in the form of words, to align with events, we use the same pre-trained language representation model to extract the vector representation\footnote{If an object word is tokenized to multiple tokens, we take their average representation as the token representation.} of selected objects and denote them as $\o_1, \o_2, ..., \o_m \in \OM$.

\subsection{Cross-Attention Module}

The purpose of the cross-attention module is to minimize the influence of noise by selecting important context objects with events and informative tokens in events with the context. Thus, the cross-attention module contains two sub-steps: (1) context representation; (2) event representation. 

\noindent \textbf{Context Representation:} For each event $e$, whose tokens' vector representations are $\w_1, \w_2, ..., \w_n$, we first take the average of all tokens and denote the resulted average vector as $\widetilde{\w}$. As the vector representation set of all selected objects is denoted as $\OM$, we compute the overall context representation as:
\begin{equation}
    \o = \sum_{\o^\prime \in \OM} a_{\widetilde{\w}, \o^\prime} \cdot \o^\prime,
\end{equation}
where $a_{\widetilde{\w}, \o^\prime}$ is the attention weight of $\widetilde{\w}$ on object $o^\prime$. Here we compute the attention weight as:
\begin{equation}
    a_{\widetilde{\w}, \o^\prime} = NN_a([\widetilde{\w},\o^\prime]),
\end{equation}
where $NN_a$ is a standard two-layer feed forward neural network and $[,]$ indicates the concatenation.

\noindent \textbf{Event Representation:} After getting the context representation, the next step is computing the event representation. Assuming that the vector set of $e$ is $\WM$, we can get the event representation with a similar attention structure:
\begin{align*}
    \e &= \sum_{\w^\prime \in \WM} b_{\o, \w^\prime} \cdot \w^\prime,\\
    b_{\o, \w^\prime} &= NN_b([\o,\w^\prime]),
\end{align*}
where $b$ is the attention weight we computed with another feed forward neural network $NN_b$.

\subsection{Causality Prediction}
Assuming that the context representations with $e_1$ and $e_2$ as attention signal are denoted as $\o_{e_1}$ and $\o_{e_2}$ respectively and the overall representations of $e_1$ and $e_2$ are $\e_1$ and $\e_2$, we can then predict the final causality score as follows:
\begin{equation}
    F(e_1,e_2,I_1,I_2) = NN_c([\e_1, \e_2, \o_{e_1}, \o_{e_2}]).
\end{equation}




\section{The Experiment}
\label{sec:experiment}

In this section, we present experiments and analysis to show that both the pre-trained textual representation and visual context can help learn causality knowledge from time-consecutive images.

\subsection{Evaluation Metric}

As each event in the first image could cause multiple events in the second image, following previous works~\cite{DBLP:conf/cvpr/XuZCF17}, we evaluate different causality extraction models with a ranking-based evaluation metric. Given each event $e$ in the first images, models are required to rank all candidate events based on how likely they think these events are caused by $e$.
We then evaluate different models based on whether the correct caused event is covered by the top one, five, or ten ranked events. We denote these evaluation metrics as Recall@1, Recall@5, and Recall@10.
In our experiment, all detected events in the same video are considered as negative examples. 

\subsection{Baseline Methods}


To prove that the context is crucial and the proposed cross-attention module is helpful, we compare VCC with the following models:
\begin{enumerate} [leftmargin=*]
    \item \textbf{No Visual Context}: Directly predicts the causal relation between events without considering the visual context. For each event, we take the average of word representations as the event representation and concatenate the representations of two events together for the final prediction. 
    \item \textbf{No Attention}: Removes the cross-attention module and uses the average word embeddings of all selected objects to represent the context. 
    \item \textbf{ResNet as Context}: Removes the object detection module and uses the average image representation extracted by ResNet-152~\cite{DBLP:conf/cvpr/HeZRS16} as the context representation. 
\end{enumerate}

Besides the aforementioned baselines, we also present the performance of a ``random guess'' baseline, which randomly ranks all candidate events and can be used as a performance lower-bound for all causality extraction models.

\begin{table*}[t]
\small
    \centering
    \begin{tabular}{l|c||c|c|c|c|c||c}
    \toprule
        Model & Metric &Sports & Socializing & Household & Personal Care & Eating & Overall \\
        \midrule
        \multirow{3}*{Random Guess} 
         & R@1 & 0.67 & 3.64 & 1.69 & 0.00 & 9.09 & 2.13\\
         & R@5 & 14.19 & 16.36 & 15.25 & 11.11 & 27.27 & 15.25\\
         & R@10 & 28.38 & 38.18 & 27.12 & 33.33 & 27.27 & 30.14\\
         
        \midrule
        \multirow{3}*{No Visual Context} 
         & R@1 & 6.76 & \textbf{9.09} & \textbf{8.47} & 0.00 & 9.09 & 7.45\\
         & R@5 & 31.08 & 30.91 & 23.73 & 22.22 & 45.45 & 29.79\\
         & R@10 & 59.46 & \textbf{63.64} & \textbf{64.41} & 55.56 & 72.73 & 61.70\\
         \midrule
        \multirow{3}*{No Attention} 
         & R@1 & 7.43 & \textbf{9.09} & 6.78 & 0.00 & 18.18 & 7.80\\
         & R@5 & \textbf{38.51} & 29.09 & 22.03 & 11.11 & 36.36 & 32.27\\
         & R@10 & 61.49 & 54.55 & 54.24 & 66.67 & 54.55 & 58.51\\
         \midrule
        \multirow{3}*{ResNet as Context} 
         & R@1 & 7.43 & \textbf{9.09} & 1.69 & 0.00 & 9.09 & 6.38\\
         & R@5 & 37.16 & 27.27 & 22.03 & 11.11 & 36.36 & 31.21\\
         & R@10 & 59.46 & 58.18 & 52.54 & \textbf{77.78} & 63.64 & 58.51\\
          \midrule
         \multirow{3}*{The proposed VCC Model} 
         & R@1 & \textbf{8.78} & 7.27 & 6.78 & \textbf{11.11} & \textbf{27.27} & \textbf{8.87}\\
         & R@5 & 37.16 & \textbf{36.36} & \textbf{28.81} & \textbf{33.33} & \textbf{45.45} & \textbf{34.75}\\
         & R@10 & \textbf{64.86} & 58.18 & 62.71 & 55.56 & \textbf{72.73} & \textbf{63.12}\\
          \bottomrule
    \end{tabular}
    \caption{Performances of different models. Performances by context categories are reported. The best performance on each evaluation metric for each category is indicated with the \textbf{bold} font.}
    \label{tab:main_result}
\end{table*}

\subsection{Implementation Details}

\noindent \textbf{Model Details:} We use BERT~\cite{DBLP:conf/naacl/DevlinCLT19} as the textual representation model to encode both events and and objects detected from images.
We follow the previous scene graph generation work~\cite{DBLP:conf/cvpr/XuZCF17} to leverage a Faster R-CNN network~\cite{DBLP:journals/pami/RenHG017}, which is trained on MS-COCO~\cite{lin2014microsoft}, to detect objects from the images. 

We set the hidden state size in the feed-forward neural network to be 200 and the number of selected objects $m$ to be 10.
The total number of trainable parameters is 109.9 million (including 109.48 million from BERT-base). 

\noindent \textbf{Training Details:} During the training phase, for each positive example, we randomly select one negative example and use cross-entropy as the loss function. We employ stochastic gradient descent (SGD) as the optimizer. All parameters are initialized randomly and the learning rate is set to be $10^{-4}$. All models are trained with up to ten epochs\footnote{All models converge before ten epochs.}, and the models that perform best on the dev set are evaluated on the test set.
The experiments are implemented on Intel(R) Xeon(R) CPU E5-2640v4@2.40GHz and one GTX-1080 GPU. Each training epoch takes 18 minutes on average.



\subsection{Result Analysis}


We report the performance of all models on all categories and show the results in Table~\ref{tab:main_result}, from which we can make the following observations: 


\begin{enumerate}[leftmargin=*]
    \item All models significantly outperform the ``random guess'' baseline in almost all settings, which show that models can learn to extract meaningful causality knowledge from these time-consecutive images and learning causality knowledge from the visual signal can be a good supplement for the current text-based approach of acquiring causality knowledge in the future.
    \item With the help of the context information, VCC outperforms the baseline `No Context' model in most experiment settings, which proves the importance of context and is consistent with our previous observation that some causality only makes sense in certain contexts. 
    \item The proposed VCC model outperforms the `No Attention' model significantly, which demonstrates the influence of noise introduced by the object detection module and the effectiveness of the proposed cross-attention module. 
    \item The proposed VCC model, which first extracts objects from images and then use extracted objects as the context, outperforms the ``ResNet as Context'' baseline, which directly uses the ResNet encoded image representation as the context, even though it does not suffer from the noise introduced by the object detection module at all. One possible explanation is that the event textual representation and the ResNet encoded image representation are vectors in different semantic spaces (one in language and one in vision), which may not perfectly align with each other. As a comparison, in the proposed model, we first extract objects from images and then encode them as text, and thus the alignment issue no longer exists. 
    \item In general, even though the proposed model outperforms all baseline methods and can learn to extract meaningful causality knowledge from the training data, the task is still challenging due to the following reasons: (1) Missing information about the visual context: the performance of current object detection module is not good enough and many important objects related to the two events could be missing; (2) Correct resolution of pronouns: pronouns appear frequently in events and without the correct resolution of those pronouns, it is hard to fully understand the semantic meaning of events; (3) Lack of support of external knowledge, especially commonsense knowledge: the proposed dataset is a great test dataset for evaluating models abilities of understanding causality but its scale is not large enough to cover all the scenarios. It is important to include more knowledge from other resources.
\end{enumerate}


\begin{table}[t]
\small
    \centering
    \begin{tabular}{l||c|c|c}
    \toprule
         & R@1  & R@5 &R@10 \\
      \midrule
      Random Guess & 2.13 & 15.25 & 30.14\\
      \midrule
      BERT & 2.13  & 22.34 & 39.00 \\
      GPT-2 & 3.55  & 17.73 & 34.40 \\
      \midrule
      VCC (BERT) & \textbf{8.87} & \textbf{34.75} & \textbf{63.12}  \\
        VCC (GPT-2) & 7.80 & 31.56 & 56.03 \\
         \bottomrule
    \end{tabular}
    \vspace{0.1in}
    \caption{Performances of different models. Best-performing models are indicated with the \textbf{bold} font.}
    \label{tab:lm}
\end{table}

\subsection{Can Language Representation Models Understand Causality?}

As observed in \cite{DBLP:conf/emnlp/PetroniRRLBWM19}, language representation models can preserve rich knowledge, 
in this subsection, we conduct experiments to investigate whether pre-trained language representation models (i.e., BERT~\cite{DBLP:conf/naacl/DevlinCLT19} and GPT-2~\cite{radford2019language}) can understand causality without any training.
For each candidate event pair (e.g., (``A dog is running'', ``The leaves blow around'')), we convert it into a natural sentence (e.g., ``A dog is running, so the leaves blow around''), and then input it into the language representation model. The overall probability returned by models can be used as their causality predictions. Higher probability indicates higher plausibility prediction. We rank all candidates by their probabilities and evaluate the models in the same way as previous experiments. 
We conduct experiments on BERT-base\footnote{We also tried BERT large, but it does not make a significant improvement over the base model.} and GPT-2 (774 million parameters).
with the Hugging face implementation\footnote{https://github.com/huggingface/transformers}.
Besides unsupervised approaches, we also present the performances of replacing the language representation module in VCC with different pre-trained models.


From the experimental results in Table~\ref{tab:lm}, we can see that, compared with the ``random guess'' baseline, unsupervised BERT and GPT-2 only achieved slightly better performance. 
The reason behind this is that even though these pre-trained language representation models contain rich semantics about events, they can only distinguish which two events are more relevant rather than identify the causality between them.
As all negative examples are also selected from the same video, which makes them very relevant to the target process, the task becomes too challenging for unsupervised models.
However, if we incorporate them into the VCC model and further train them, they will achieve much better performance, which shows that if we allow further training, the model will learn how to better use the contained rich semantics and thus achieve better performance.

\subsection{Case Study}

\begin{figure}
    \centering
    \includegraphics[width=0.6\linewidth]{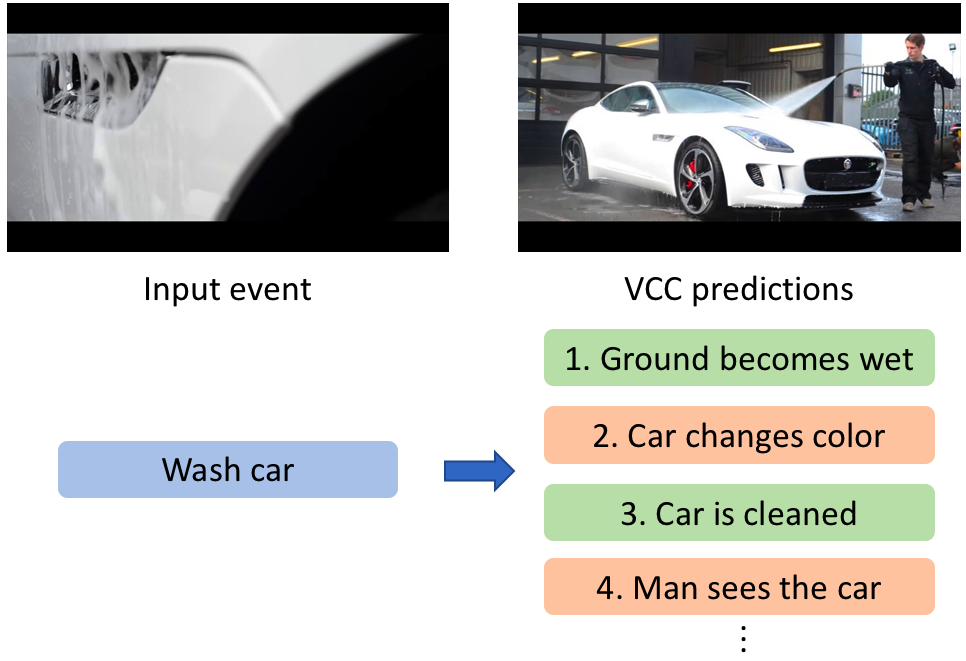}
    \caption{Given an event `wash car' in the first image, we show the top four events that VCC predicts to be contained in the second image and caused by `wash car'. Correct and wrong predictions are indicated with green and orange background respectively.}
    \label{fig:case-study}
\end{figure}

To further analyze the success and limitation of the proposed VCC model, we present a case study, where predictions are sorted based the prediction scores, in Figure~\ref{fig:case-study}, from which we can see that VCC successfully predict that ``wash car'' can cause ``ground becomes wet'' and ``car is cleaned'', but it also makes mistakes.
For example, based on these two images, humans know that the car does not change color but VCC may mistakenly connect the event ``change color'' with ``wash'' from some other training examples.
In this case, using a few objects to represent images may not be enough to cover all the visual information.
Besides that, VCC also predicts ``Man sees the car'', which indeed happens in the second image but is not caused by ``wash''.
Understanding this might need the inference over external knowledge.
How to leverage external knowledge to better understand the visual signal and thus acquire more accurate causal knowledge is left for our future investigation.



\section{Related Works}\label{sec:related-work}
In this section, we introduce related work about causality acquisition and visually-grounded NLP.
\subsection{Causality Acquisition}
As a crucial knowledge for many artificial intelligence (AI) systems~\cite{pearl2018book}, causality has long been an important research topic in many communities with different focuses.
For example, in the machine learning community, researchers~\cite{pearl2018book,schlkopf2019causality} are focusing on modeling causality from structured data (e.g., directed acyclic graph).
Different from them, researchers from the computer vision community~\cite{DBLP:journals/tist/FireZ16,DBLP:conf/cvpr/EhsaniBRMF18} are focusing on identifying key objects or events in images that can cause certain decision makings.
Last but not least, previous works in the natural language processing (NLP) community are mostly working on acquiring causality knowledge via either crowd-sourcing~\cite{liu2004conceptnet,DBLP:conf/aaai/SapBABLRRSC19} or linguistic pattern mining~\cite{DBLP:conf/acl/HideyM16} and then applying the acquired knowledge for understanding human language~\cite{DBLP:conf/acl/RothWNF18}.
The ultimate goal of this paper is the same as previous NLP works that we are trying to acquire causality knowledge, which can be stored and used for downstream tasks.
But the approach is different. To the best of our knowledge, this is the first work exploring the possibility of directly acquiring causality from the visual signal.
Another related work from the CV community is Visual-COPA~\cite{DBLP:conf/lrec/YeoLWCCAH18}, which asks models to identify if one image can cause another. The major difference is that our paper is trying to extract causality knowledge rather than leveraging external knowledge to predict image relations.


\subsection{Visually-grounded NLP}

As the intersection of computer vision (CV) and natural language processing (NLP),
visually-grounded NLP research topics are popular in both communities.
For example, image captioning~\cite{DBLP:conf/icml/XuBKCCSZB15} aims at generating captions for images and scene graph generation~\cite{DBLP:conf/cvpr/XuZCF17} tries to detect not just entities but also events or states from images.
Besides these basic tasks, some other visually-grounded NLP tasks (e.g., visual question answering (VQA)~\cite{DBLP:conf/iccv/AntolALMBZP15} and visual dialogue~\cite{DBLP:conf/cvpr/DasKGSYMPB17}) are also created to test how well models can understand human language and visual signals jointly.
Another line of related works is visual commonsense reasoning~\cite{DBLP:conf/naacl/YatskarOF16,DBLP:conf/emnlp/YangGSC18,DBLP:conf/cvpr/ZellersBFC19,DBLP:conf/eccv/ParkBMFC20}, which aim at either extracting commonsense knowledge from the images or evaluating models' commonsense reasoning abilities over images.
Considering that the causality often happens between events that appear in the temporal order, which are unlikely to appear in the same image, we choose to work on time-consecutive image pairs rather than a single image.

\section{Conclusion}\label{sec:conclusion}
In this paper, we explore the possibility of learning causality knowledge from time-consecutive images.
To do so, we first formally define the task and then create a high-quality dataset Vis-Causal
, which contains 4,000 image pairs, 23,558 event pairs, and causal relation annotations under two settings.
On top of the collected dataset, we propose a Vision-Contextual Causal (VCC) model to demonstrate that with the help of strong pre-trained textual and visual representations and careful training, it is possible to directly acquire contextual causality from visual signals.
Further analysis shows that even though VCC can outperform all baseline methods, it is still not perfect.
As the visual signal could serve as an important causality knowledge resource, we will keep exploring how to better acquire causal knowledge from the visual signal (e.g., leveraging external knowledge) in the future. 
Both the dataset and code will be released to encourage research on the causality acquisition.

\section*{Acknowledgements}
This paper was supported by  Early Career Scheme (ECS, No. 26206717), General Research Fund (GRF, No. 16211520), and Research Impact Fund (RIF, No. R6020-19) from the Research Grants Council (RGC) of Hong Kong. 
This research was also supported by the Office of the Director of National Intelligence (ODNI), Intelligence Advanced Research Projects Activity (IARPA), via IARPA Contract No. 2019-19051600006 under the BETTER Program, and by contract FA8750-19-2-1004 with the US Defense Advanced Research Projects Agency (DARPA). The views expressed are those of the authors and do not reflect the official policy or position of the Department of Defense or the U.S. Government.


\bibliographystyle{unsrt}  
\bibliography{Visual-commonsense}  

\begin{thebibliography}{10}

\bibitem{DBLP:conf/acl/OhTHSSO13}
Jong{-}Hoon Oh, Kentaro Torisawa, Chikara Hashimoto, Motoki Sano, Stijn~De
  Saeger, and Kiyonori Ohtake.
\newblock Why-question answering using intra- and inter-sentential causal
  relations.
\newblock In {\em Proceedings of ACL 2013}, pages 1733--1743, 2013.

\bibitem{DBLP:conf/acl/HashimotoTKSVOK14}
Chikara Hashimoto, Kentaro Torisawa, Julien Kloetzer, Motoki Sano, Istv{\'{a}}n
  Varga, Jong{-}Hoon Oh, and Yutaka Kidawara.
\newblock Toward future scenario generation: Extracting event causality
  exploiting semantic relation, context, and association features.
\newblock In {\em Proceedings of ACL 2014}, pages 987--997, 2014.

\bibitem{DBLP:conf/acl/RothWNF18}
Qiang Ning, Zhili Feng, Hao Wu, and Dan Roth.
\newblock Joint reasoning for temporal and causal relations.
\newblock In {\em Proceedings of ACL 2018}, pages 2278--2288, 2018.

\bibitem{pearl2018book}
Judea Pearl and Dana Mackenzie.
\newblock {\em The book of why: the new science of cause and effect}.
\newblock Basic Books, 2018.

\bibitem{bunge2017causality}
Mario Bunge.
\newblock {\em Causality and modern science}.
\newblock Routledge, 2017.

\bibitem{liu2004conceptnet}
Hugo Liu and Push Singh.
\newblock Conceptnet—a practical commonsense reasoning tool-kit.
\newblock {\em BT technology journal}, 22(4):211--226, 2004.

\bibitem{DBLP:conf/aaai/SapBABLRRSC19}
Maarten Sap, Ronan~Le Bras, Emily Allaway, Chandra Bhagavatula, Nicholas
  Lourie, Hannah Rashkin, Brendan Roof, Noah~A. Smith, and Yejin Choi.
\newblock {ATOMIC:} an atlas of machine commonsense for if-then reasoning.
\newblock In {\em Proceedings of AAAI 2019}, pages 3027--3035, 2019.

\bibitem{DBLP:conf/acl/HideyM16}
Christopher Hidey and Kathy McKeown.
\newblock Identifying causal relations using parallel wikipedia articles.
\newblock In {\em Proceedings of ACL 2016}, 2016.

\bibitem{DBLP:conf/kr/LuoSZHW16}
Zhiyi Luo, Yuchen Sha, Kenny~Q. Zhu, Seung{-}won Hwang, and Zhongyuan Wang.
\newblock Commonsense causal reasoning between short texts.
\newblock In {\em Proceedings of KR 2016}, pages 421--431, 2016.

\bibitem{ASER2020}
Hongming Zhang, Xin Liu, Haojie Pan, Yangqiu Song, and Cane Wing-Ki Leung.
\newblock {ASER}: A large-scale eventuality knowledge graph.
\newblock In {\em Proceedings of WWW 2020}, pages 201--211, 2020.

\bibitem{DBLP:conf/cvpr/XuZCF17}
Danfei Xu, Yuke Zhu, Christopher~B. Choy, and Li~Fei{-}Fei.
\newblock Scene graph generation by iterative message passing.
\newblock In {\em Proceedings of CVPR 2017}, pages 3097--3106, 2017.

\bibitem{DBLP:conf/eccv/YangLLBP18}
Jianwei Yang, Jiasen Lu, Stefan Lee, Dhruv Batra, and Devi Parikh.
\newblock Graph {R-CNN} for scene graph generation.
\newblock In {\em Proceedings of ECCV 2018}, pages 690--706, 2018.

\bibitem{DBLP:conf/eccv/LiOZSZW18}
Yikang Li, Wanli Ouyang, Bolei Zhou, Jianping Shi, Chao Zhang, and Xiaogang
  Wang.
\newblock Factorizable net: An efficient subgraph-based framework for scene
  graph generation.
\newblock In {\em Proceedings of ECCV 2018}, pages 346--363, 2018.

\bibitem{DBLP:conf/cvpr/HeilbronEGN15}
Fabian~Caba Heilbron, Victor Escorcia, Bernard Ghanem, and Juan~Carlos Niebles.
\newblock Activitynet: {A} large-scale video benchmark for human activity
  understanding.
\newblock In {\em Proceedings of CVPR 2015}, pages 961--970, 2015.

\bibitem{DBLP:conf/emnlp/ReisingerM10}
Joseph Reisinger and Raymond~J. Mooney.
\newblock A mixture model with sharing for lexical semantics.
\newblock In {\em Proceedings of EMNLP 2010}, pages 1173--1182, 2010.

\bibitem{DBLP:journals/coling/HillRK15}
Felix Hill, Roi Reichart, and Anna Korhonen.
\newblock Simlex-999: Evaluating semantic models with (genuine) similarity
  estimation.
\newblock {\em Computational Linguistics}, 41(4):665--695, 2015.

\bibitem{DBLP:conf/nips/VaswaniSPUJGKP17}
Ashish Vaswani, Noam Shazeer, Niki Parmar, Jakob Uszkoreit, Llion Jones,
  Aidan~N. Gomez, Lukasz Kaiser, and Illia Polosukhin.
\newblock Attention is all you need.
\newblock In {\em Proceedings of NIPS 2017}, pages 5998--6008, 2017.

\bibitem{DBLP:conf/naacl/DevlinCLT19}
Jacob Devlin, Ming{-}Wei Chang, Kenton Lee, and Kristina Toutanova.
\newblock {BERT:} pre-training of deep bidirectional transformers for language
  understanding.
\newblock In {\em Proceedings of NAACL-HLT 2019}, pages 4171--4186, 2019.

\bibitem{DBLP:conf/cvpr/DasKGSYMPB17}
Abhishek Das, Satwik Kottur, Khushi Gupta, Avi Singh, Deshraj Yadav, Jos{\'{e}}
  M.~F. Moura, Devi Parikh, and Dhruv Batra.
\newblock Visual dialog.
\newblock In {\em Proceedings of CVPR 2017}, pages 1080--1089, 2017.

\bibitem{DBLP:conf/cvpr/HeZRS16}
Kaiming He, Xiangyu Zhang, Shaoqing Ren, and Jian Sun.
\newblock Deep residual learning for image recognition.
\newblock In {\em Proceedings of CVPR 2016}, pages 770--778, 2016.

\bibitem{DBLP:journals/pami/RenHG017}
Shaoqing Ren, Kaiming He, Ross~B. Girshick, and Jian Sun.
\newblock Faster {R-CNN:} towards real-time object detection with region
  proposal networks.
\newblock {\em {IEEE} Trans. Pattern Anal. Mach. Intell.}, 39(6):1137--1149,
  2017.

\bibitem{lin2014microsoft}
Tsung-Yi Lin, Michael Maire, Serge Belongie, James Hays, Pietro Perona, Deva
  Ramanan, Piotr Doll{\'a}r, and C~Lawrence Zitnick.
\newblock Microsoft coco: Common objects in context.
\newblock In {\em European conference on computer vision}, pages 740--755.
  Springer, 2014.

\bibitem{DBLP:conf/emnlp/PetroniRRLBWM19}
Fabio Petroni, Tim Rockt{\"{a}}schel, Sebastian Riedel, Patrick S.~H. Lewis,
  Anton Bakhtin, Yuxiang Wu, and Alexander~H. Miller.
\newblock Language models as knowledge bases?
\newblock In {\em Proceedings of EMNLP-IJCNLP 2019}, pages 2463--2473, 2019.

\bibitem{radford2019language}
Alec Radford, Jeff Wu, Rewon Child, David Luan, Dario Amodei, and Ilya
  Sutskever.
\newblock Language models are unsupervised multitask learners.
\newblock {\em OpenAI Blog}, 1(8):9, 2019.

\bibitem{schlkopf2019causality}
Bernhard Schölkopf.
\newblock Causality for machine learning, 2019.

\bibitem{DBLP:journals/tist/FireZ16}
Amy~Sue Fire and Song{-}Chun Zhu.
\newblock Learning perceptual causality from video.
\newblock {\em {ACM} {TIST}}, 7(2):23:1--23:22, 2016.

\bibitem{DBLP:conf/cvpr/EhsaniBRMF18}
Kiana Ehsani, Hessam Bagherinezhad, Joseph Redmon, Roozbeh Mottaghi, and Ali
  Farhadi.
\newblock Who let the dogs out? modeling dog behavior from visual data.
\newblock In {\em Proceedings of CVPR 2018}, pages 4051--4060, 2018.

\bibitem{DBLP:conf/lrec/YeoLWCCAH18}
Jinyoung Yeo, Gyeongbok Lee, Gengyu Wang, Seungtaek Choi, Hyunsouk Cho,
  Reinald~Kim Amplayo, and Seung{-}won Hwang.
\newblock Visual choice of plausible alternatives: An evaluation of image-based
  commonsense causal reasoning.
\newblock In {\em Proceedings of LREC 2018}, 2018.

\bibitem{DBLP:conf/icml/XuBKCCSZB15}
Kelvin Xu, Jimmy Ba, Ryan Kiros, Kyunghyun Cho, Aaron~C. Courville, Ruslan
  Salakhutdinov, Richard~S. Zemel, and Yoshua Bengio.
\newblock Show, attend and tell: Neural image caption generation with visual
  attention.
\newblock In {\em Proceedings of ICML 2015}, pages 2048--2057, 2015.

\bibitem{DBLP:conf/iccv/AntolALMBZP15}
Stanislaw Antol, Aishwarya Agrawal, Jiasen Lu, Margaret Mitchell, Dhruv Batra,
  C.~Lawrence Zitnick, and Devi Parikh.
\newblock {VQA:} visual question answering.
\newblock In {\em Proceedings of ICCV 2015}, pages 2425--2433, 2015.

\bibitem{DBLP:conf/naacl/YatskarOF16}
Mark Yatskar, Vicente Ordonez, and Ali Farhadi.
\newblock Stating the obvious: Extracting visual common sense knowledge.
\newblock In {\em Proceedings of NAACL-HLT 2016}, pages 193--198, 2016.

\bibitem{DBLP:conf/emnlp/YangGSC18}
Shaohua Yang, Qiaozi Gao, Sari Saba{-}Sadiya, and Joyce~Yue Chai.
\newblock Commonsense justification for action explanation.
\newblock In {\em Proceedings of EMNLP 2018}, pages 2627--2637, 2018.

\bibitem{DBLP:conf/cvpr/ZellersBFC19}
Rowan Zellers, Yonatan Bisk, Ali Farhadi, and Yejin Choi.
\newblock From recognition to cognition: Visual commonsense reasoning.
\newblock In {\em Proceedings of CVPR 2019}, pages 6720--6731, 2019.

\bibitem{DBLP:conf/eccv/ParkBMFC20}
Jae~Sung Park, Chandra Bhagavatula, Roozbeh Mottaghi, Ali Farhadi, and Yejin
  Choi.
\newblock Visualcomet: Reasoning about the dynamic context of a still image.
\newblock In {\em Proceedings of ECCV 2020}, pages 508--524, 2020.

\end{thebibliography}

\clearpage

\appendix

\section{Appendix}

\begin{figure*}[h]
    \centering
    \includegraphics[width=0.6\linewidth]{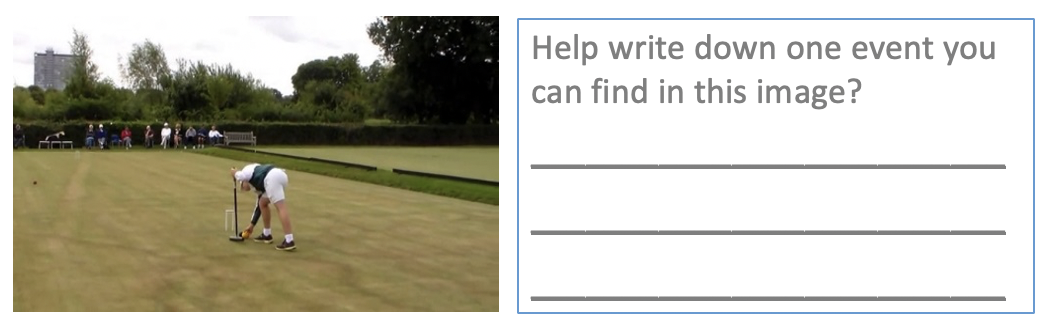}
    \caption{Event identification. Annotators are requested to identify one event from the image. Detailed instructions and examples, which are not shown in this screenshot, are also provided at the beginning of each survey.}
    \label{fig:eventuality-identification}
\end{figure*}

\begin{figure*}[h]
    \centering
    \includegraphics[width=0.6\linewidth]{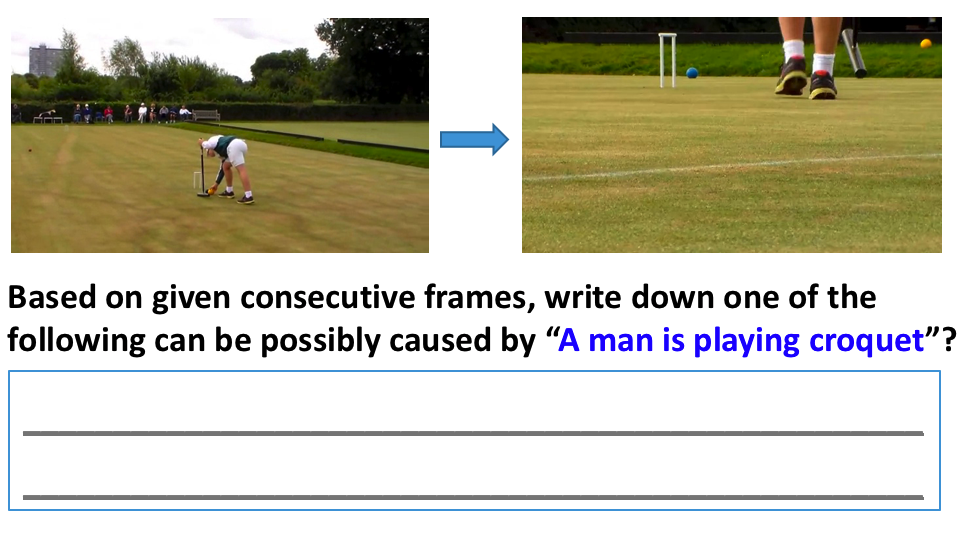}
    \caption{First-round Causal Relation Annotation. Annotators are requested to identify events happen in the second image that can be caused by the event in the first image.}
    \label{fig:first-step}
\end{figure*}

\begin{figure*}[h]
    \centering
    \includegraphics[width=0.5\linewidth]{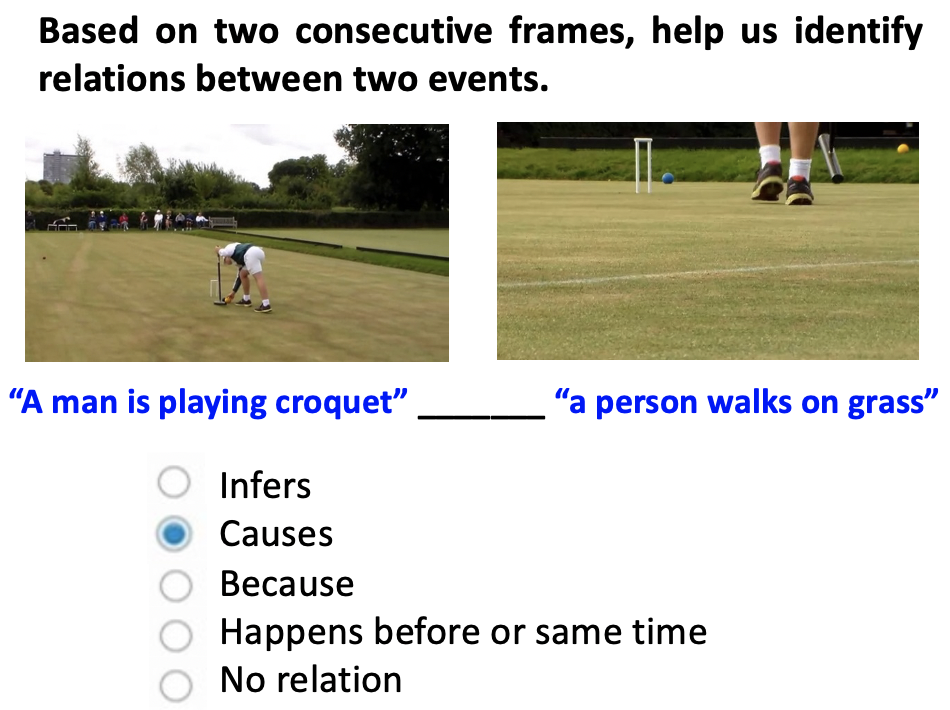}
    \caption{Second-round Causal Relation Annotation. We only show the survey for causal relation annotation with context as an example. The only difference in the w/o context setting is that we remove the images.}
    \label{fig:second-step}
\end{figure*}

\end{document}